\documentclass[11pt,a4paper]{article}

\usepackage{authblk}
\usepackage[hyphens]{url}
\usepackage[hyperref]{acl2018} 

\usepackage{times}
\usepackage{latexsym}
\usepackage{amsmath,amsfonts,amssymb}

\usepackage{graphicx}
\usepackage{algorithm}
\usepackage[noend]{algorithmic}

\usepackage{booktabs}
\usepackage{todonotes}

\aclfinalcopy
\def\aclpaperid{1262}

\newcommand{\REQUIREP}{\item[\hphantom{\textbf{Input:}}]}

\DeclareMathOperator{\NN}{\mathrm{NN}}

\DeclareMathOperator{\nmpu}{\mathrm{nmPU}}
\DeclareMathOperator{\nipu}{\mathrm{niPU}}

\newcommand{\watset}{\textsc{Watset}}

\newcommand\legend[1]{\fcolorbox{white}{#1}{\rule{0pt}{6pt}\rule{6pt}{0pt}}}
\definecolor{ggplotverb}{HTML}{b2abd2}
\definecolor{ggplotsubject}{HTML}{e66101}
\definecolor{ggplotobject}{HTML}{fdb863}
\definecolor{ggplotframe}{HTML}{5e3c99}

\usepackage{array}
\newcolumntype{R}[1]{>{\raggedleft\let\newline\\\arraybackslash\hspace{0pt}}m{#1}}

\title{Unsupervised Semantic Frame Induction using Triclustering}

\author[$\dag$]{\textbf{Dmitry Ustalov}}
\author[$\ddag$]{\textbf{Alexander Panchenko}}
\author[$\star$]{\textbf{Andrei Kutuzov}}
\author[$\ddag$]{\\\textbf{Chris Biemann}}
\author[$\dag$]{\textbf{Simone Paolo Ponzetto}}

\affil[$\dag$]{University of Mannheim, Germany}
\affil[ ]{\texttt{\{dmitry,simone\}@informatik.uni-mannheim.de}}

\affil[$\star$]{University of Oslo, Norway}
\affil[ ]{\texttt{andreku@ifi.uio.no}}

\affil[$\ddag$]{University of Hamburg, Germany}
\affil[ ]{\texttt{\{panchenko,biemann\}@informatik.uni-hamburg.de}}

\date{}

\begin{document}

\setlength{\titlebox}{55mm}
\setlength{\floatsep}{0pt}
\setlength{\abovecaptionskip}{4pt}
\setlength{\belowcaptionskip}{0pt}
\setlength{\abovetopsep}{0pt}
\setlength{\belowbottomsep}{0pt}
\setlength{\aboverulesep}{0pt}
\setlength{\belowrulesep}{0pt}

\maketitle

\begin{abstract}\vspace{-.5em}
We use dependency triples automatically extracted from a Web-scale corpus to perform unsupervised semantic frame induction. We cast the frame induction problem as a \textit{triclustering} problem that is a generalization of clustering for \textit{triadic} data. Our replicable benchmarks demonstrate that the proposed graph-based approach, \textit{Triframes}, shows state-of-the art results on this task on a FrameNet-derived dataset and performing on par with competitive methods on a verb class clustering task.
\end{abstract}

\section{Introduction}

Recent years have seen much work on Frame Semantics \cite{Fillmore:82}, enabled by the availability of a large set of frame definitions, as well as a manually annotated text corpus provided by the FrameNet project \cite{Baker:98}. FrameNet data enabled the development of wide-coverage frame parsers using supervised learning \cite[\emph{inter alia}]{Gildea:02,Erk:06,Das:14}, as well as its application to a wide range of tasks, ranging from  answer extraction in Question Answering \cite{Shen:07} and Textual Entailment \cite{Burchardt:09,BenAharon:10}.

However, frame-semantic resources are arguably expensive and time-consuming to build due to difficulties in defining the frames, their granularity and domain, as well as the complexity of the construction and annotation tasks requiring expertise in the underlying knowledge. Consequently, such resources exist only for a few languages~\cite{Boas:09} 
and even English is lacking domain-specific frame-based resources. Possible inroads are cross-lingual semantic annotation transfer \cite{Pado:09,Hartmann:16} or linking FrameNet to other lexical-semantic or ontological resources  \cite[\emph{inter alia}]{Narayanan:03,Tonelli:09,Laparra:10,Gurevych:12}. But while the arguably simpler task of PropBank-based Semantic Role Labeling has been successfully addressed by unsupervised approaches \cite{Lang:10,Titov:11}, fully unsupervised frame-based semantic annotation exhibits far more challenges, starting with the preliminary step of automatically inducing a set of semantic frame definitions that would drive a subsequent text annotation. In this work, we aim at overcoming these issues by automatizing the process of FrameNet construction through unsupervised frame induction techniques.

\begin{table}[t]
\centering
\footnotesize
\begin{tabular}{lll}
\textbf{FrameNet~~~~~~~} & \textbf{Role~~~~~~~~} & \textbf{Lexical Units (LU)} \\\toprule
\textit{Perpetrator} & Subject & kidnapper, alien, militant \\ \midrule
\textit{FEE}         & Verb    & snatch, kidnap, abduct \\ \midrule
\textit{Victim}      & Object  & son, people, soldier, child \\
\end{tabular}
\caption{\label{tab:tricluster}Example of a LU tricluster corresponding to the ``Kidnapping'' frame from FrameNet.}
\end{table}

\paragraph{Triclustering.} In this work, we cast the frame induction problem as a \textit{triclustering} task~\cite{Zhao:05,Ignatov:15}, namely a generalization of standard clustering and bi-clustering~\cite{Cheng:00}, aiming at simultaneously clustering objects along three dimensions (cf. Table~\ref{tab:tricluster}). First, using triclustering allows to avoid sequential nature of frame induction approaches, e.g. \cite{Kawahara:14}, where two independent clusterings are needed. Second, benchmarking frame induction as triclustering against other methods on dependency triples allows to abstract away the evaluation of the frame induction algorithm from other factors, e.g., the input corpus or pre-processing steps, thus allowing a fair comparison of different induction models. 

The \textbf{contributions of this paper} are three-fold: (1) we are the \textit{first to apply} triclustering algorithms for unsupervised frame induction, (2) we propose a \textit{new approach to triclustering}, achieving state-of-the-art performance on the frame induction task, (3) we propose a \textit{new method for the evaluation} of frame induction enabling straightforward comparison of approaches. In this paper, we focus on the simplest setup with \textit{subject-verb-object} (SVO) triples and two roles, but our evaluation framework can be extended to more roles.

In contrast to the recent approaches like the one by \citet{Jauhar:17}, our approach induces semantic frames without any supervision, yet capturing only two core roles: the subject and the object of a frame triggered by verbal predicates. Note that it is not generally correct to expect that the SVO triples obtained by a dependency parser are necessarily the core arguments of a predicate. Such roles can be implicit, i.e., unexpressed in a given context \cite{Schenk:16}. Keeping this limitation in mind, we assume that the triples obtained from a Web-scale corpus cover most core arguments sufficiently.


\paragraph{Related Work.} \textit{LDA-Frames}~\cite{Materna:12,Materna:13} is an approach to inducing semantic frames using LDA~\cite{Blei:03} for generating semantic frames and their respective frame-specific semantic roles at the same time. The authors evaluated their approach against the CPA corpus~\cite{Hanks:05}. \textit{ProFinder}~\cite{Cheung:13} is another generative approach that also models both frames and roles as latent topics. The evaluation was performed on the in-domain information extraction task MUC-4~\cite{Sundheim:92} and on the text summarization task TAC-2010.\footnote{\url{https://tac.nist.gov/2010/Summarization}} \citet{Modi:12} build on top of an unsupervised semantic role labeling model~\cite{Titov:12}. The raw text of sentences from the FrameNet data is used for training. The FrameNet gold annotations are then used to evaluate the labeling of the obtained frames and roles, effectively clustering instances known during induction. \citet{Kawahara:14} harvest a huge collection of verbal predicates along with their argument instances and then apply the Chinese Restaurant Process clustering algorithm to group predicates with similar arguments. The approach was evaluated on the verb cluster dataset of \citet{Korhonen:03}.

A major issue with unsupervised frame induction task is that these and some other related approaches, e.g., \cite{OConnor:13}, were all evaluated in completely different incomparable settings, and used different input corpora. In this paper, we propose a methodology to resolve this issue.

\section{The Triframes Algorithm}

Our approach to frame induction relies on graph clustering. We focused on a simple setup using two roles and the SVO triples, arguing that it still can be useful, as frame roles are primarily expressed by subjects and objects, giving rise to semantic structures extracted in an unsupervised way with high coverage.

\paragraph{Input Data.} As the input data, we use SVO triples extracted by a dependency parser. According to our statistics on the dependency-parsed FrameNet corpus of over 150 thousand sentences~\cite{Bauer:12}, the SUBJ and OBJ relationships are the two most common shortest paths between frame evoking elements (\textit{FEEs}) and their roles, accounting for 13.5~\% of instances of a heavy-tail distribution of over 11 thousand different paths that occur three times or more in the FrameNet data. While this might seem a simplification that does not cover prepositional phrases and frames filling the roles of other frames in a nested fashion, we argue that the overall frame inventory can be induced on the basis of this restricted set of constructions, leaving other paths and more complex instances for further work.

\paragraph{The Method.} Our method constructs embeddings for SVO triples to reduce the frame induction problem to a simpler graph clustering problem. Given the vocabulary $V$, a $d$-dimensional word embedding model $v \in V \rightarrow \vec{v} \in \mathbb{R}^d$, and a set of SVO triples $T \subseteq V^3$ extracted from a syntactically analyzed corpus, we construct the triple similarity graph $\mathcal{G}$. Clustering of $\mathcal{G}$ yields sets of triples corresponding to the instances of the semantic frames, thereby clustering frame-evoking predicates and roles simultaneously.

We obtain dense representations of the triples $T$ by concatenating the word vectors corresponding to the elements of each triple by transforming a triple $t = (s, p, o) \in T$ into the $(3d)$-dimensional vector $\vec{t} = \vec{s} \oplus \vec{p} \oplus \vec{o}$. Subsequently, we use the triple embeddings to generate the undirected graph $\mathcal{G} = (T, E)$ by constructing the edge set $E \subseteq T^2$. For that, we compute $k \in \mathbb{N}$ nearest neighbors of each triple vector $\vec{t} \in \mathbb{R}^{3d}$ and establish cosine similarity-weighted edges between the corresponding triples.

Then, we assume that the triples representing similar contexts appear in similar roles, which is explicitly encoded by the concatenation of the corresponding vectors of the words constituting the triple. We use graph clustering of $\mathcal{G}$ to retrieve communities of similar triples forming frame clusters; a clustering algorithm is a function $\textsc{Cluster} : (T, E) \rightarrow \mathbb{C}$ such that $T = \bigcup_{C \in \mathbb{C}} C$. Finally, for each cluster $C \in \mathbb{C}$, we aggregate the subjects, the verbs, and the objects of the contained triples into separate sets. As the result, each cluster is transformed into a \textit{triframe}, which is a triple that is composed of the subjects $f_s \subseteq V$, the verbs $f_v \subseteq V$, and the objects $f_o \subseteq V$.

\begin{algorithm}[t]
\caption{\textit{Triframes} frame induction}
\label{alg:triframe}
\begin{algorithmic}[1]
\REQUIRE{an embedding model $v \in V \rightarrow \vec{v} \in \mathbb{R}^d$,}
\REQUIREP{a set of SVO triples $T \subseteq V^3$,}
\REQUIREP{the number of nearest neighbors $k \in \mathbb{N}$,}
\REQUIREP{a graph clustering algorithm $\textsc{Cluster}$.}
\ENSURE{a set of triframes $F$.}
\STATE{$S \gets \{t \!\rightarrow \vec{t} \in \mathbb{R}^{3d} : t \in T\}$}
\STATE{$E \gets \{(t, t') \in T^2 : t' \in \NN^S_k(\vec{t}), t \neq t'\}$}
\STATE{$F \gets \emptyset$}
\FORALL{$C \in \textsc{Cluster}(T, E)$}
\STATE{$f_s \gets \{s \in V : (s, v, o) \in C\}$}
\STATE{$f_v \gets \{v \in V : (s, v, o) \in C\}$}
\STATE{$f_o \gets \{o \in V : (s, v, o) \in C\}$}
\STATE{$F \gets F \cup \{(f_s, f_v, f_o)\}$}
\ENDFOR
\RETURN{$F$}
\end{algorithmic}
\end{algorithm}

Our frame induction approach outputs a set of triframes $F$ as presented in Algorithm~\ref{alg:triframe}. The hyper-parameters of the algorithm are the number of nearest neighbors for establishing edges ($k$) and the graph clustering algorithm $\textsc{Cluster}$. During the concatenation of the vectors for words forming triples, the $(|T| \times 3d)$-dimensional vector space $S$ is created. Thus, given the triple $t \in T$, we denote the $k$ nearest neighbors extraction procedure of its concatenated embedding from $S$ as $\NN^S_k(\vec{t}) \subseteq T$. We used $k=10$ nearest neighbors per triple.

To cluster the nearest neighbor graph of SVO triples $\mathcal{G}$, we use the {\watset} \textit{fuzzy graph clustering} algorithm~\cite{Ustalov:17:acl}. It treats the vertices $T$ of the input graph $\mathcal{G}$ as the SVO triples, induces their senses, and constructs an intermediate sense-aware representation that is clustered using the Chinese Whispers (CW) hard clustering algorithm~\cite{Biemann:06}. We chose {\watset} due to its performance on the related synset induction task, its fuzzy nature, and the ability to find the number of frames automatically.

\section{Evaluation}\label{sec:evaluation}

\paragraph{Input Corpus.} 
In our evaluation, we use triple frequencies from the DepCC dataset~\cite{Panchenko:18:depcc} , which is a dependency-parsed version of the Common Crawl corpus, and the standard 300-dimensional word embeddings model trained on the Google News corpus~\cite{Mikolov:13}. All evaluated algorithms are executed on the same set of triples, eliminating variations due to different corpora or pre-processing.

\paragraph{Datasets.} We cast the complex multi-stage frame induction task as a straightforward triple clustering task. We constructed a gold standard set of triclusters, each corresponding to a FrameNet frame, similarly to the one illustrated in Table~\ref{tab:tricluster}. To construct the evaluation dataset, we extracted frame annotations from the over 150 thousand sentences from the FrameNet 1.7~\cite{Baker:98}. Each sentence contains data about the frame, FEE, and its arguments, which were used to generate triples in the form $(\text{word}_i\colon \textit{role}_1, \text{word}_j\colon \textit{FEE}, \text{word}_k\colon \textit{role}_2)$, where $\text{word}_{i/j/k}$ correspond to the roles and FEE in the sentence. We omitted roles expressed by multiple words as we use dependency parses, where one node represents a single word only.

For the sentences where more than two roles are present, all possible triples were generated. Sentences with less than two roles were omitted. Finally, for each frame, we selected only two roles, which are most frequently co-occurring in the FrameNet annotated texts. This has left us with about 100 thousand instances for the evaluation. For the evaluation purposes, we operate on the intersection of triples from DepCC and FrameNet. Experimenting on the full set of DepCC triples is only possible for several methods that scale well ({\watset}, CW, $k$-means), but is prohibitively expensive for other methods (LDA-Frames, NOAC).

In addition to the frame induction evaluation, where subjects, objects, and verbs are evaluated together, we also used a dataset of polysemous verb classes introduced in~\cite{Korhonen:03} and employed by~\citet{Kawahara:14}. Statistics of both datasets are summarized in Table~\ref{tab:datasets}. Note that the polysemous verb dataset is rather small, whereas the FrameNet triples set is fairly large, enabling reliable comparisons.

\begin{table}[t]
\centering
\resizebox{\linewidth}{!}{
\begin{tabular}{lrrr}
\textbf{Dataset} & \textbf{\# instances} & \textbf{\# unique} & \textbf{\# clusters} \\\toprule
FrameNet Triples   & 99,744 & 94,170 & 383 \\
Poly. Verb Classes &    246 &    110 & 62 \\
\end{tabular}
}
\caption{\label{tab:datasets}Statistics of the evaluation datasets.}
\end{table}

\paragraph{Evaluation Measures.} Following the approach for verb class evaluation by \citet{Kawahara:14}, we employ \textit{normalized modified purity} ($\nmpu$) and \textit{normalized inverse purity} ($\nipu$) as the clustering quality measures. Given the set of the obtained clusters $K$ and the set of the gold clusters $G$, normalized modified purity quantifies the clustering precision as the average of the weighted overlap $\delta_{K_i}(K_i \cap G_j)$ between each cluster $K_i \in K$ and the gold cluster $G_j \in G$ that maximizes the overlap with $K_i$: {\small${\nmpu = \frac{1}{N} \sum^{|K|}_{i \mbox{ s.t. } |K_i| > 1} \max_{1 \leq j \leq |G|} \delta_{K_i}(K_i \cap G_j)}$}, where the weighted overlap is the sum of the weights $c_{iv}$ for each word $v$ in $i$-th cluster: {\small${\delta_{K_i}(K_i \cap G_j) = \sum_{v \in K_i \cap G_j} c_{iv}}$}. Note that $\nmpu$ counts all the singleton clusters as wrong. Similarly, normalized inverse purity (collocation) quantifies the clustering recall: {\small${\nipu = \frac{1}{N} \sum^{|G|}_{j = 1} \max_{1 \leq i \leq |K|} \delta_{G_j}(K_i \cap G_j)}$}. $\nmpu$ and $\nipu$ are combined together as the harmonic mean to yield the overall clustering F-score ($\mathrm{F}_1$), which we use to rank the approaches.

Our framework can be extended to evaluation of more than two roles by generating more roles per frame. Currently, given a set of gold triples generated from the FrameNet, each triple element has a role, e.g., ``\textit{Victim}'', ``\textit{Predator}'', and ``\textit{FEE}''. We use fuzzy clustering evaluation measure which operates not on triples, but instead on a  set of tuples. Consider for instance a gold triple {\small$(\text{Freddy}\colon \textit{Predator}, \text{kidnap}\colon \textit{FEE}, \text{kid}\colon \textit{Victim})$}. It will be converted to three pairs {\small$(\text{Freddy}, \textit{Predator})$}, {\small$(\text{kidnap}, \textit{FEE})$}, {\small$(\text{kid}, \textit{Victim})$}. Each cluster in both $K$ and $G$ is transformed into a union of all constituent typed pairs. The quality measures are finally calculated between these two sets of tuples, $K$, and $G$. Note that one can easily pull in more than two core roles by adding to this gold standard set of tuples other roles of the frame, e.g., {\small$(\text{forest}, \textit{Location})$}. In our experiments, we focused on two main roles as our contribution is related to the application of triclustering methods. However, if more advanced methods of clustering are used, yielding clusters of arbitrary modality ($n$-clustering), one could also use our evaluation schema.

\paragraph{Baselines.} We compare our method to several available state-of-the-art baselines applicable to our dataset of triples.

\textit{LDA-Frames} by \citet{Materna:12,Materna:13} is a frame induction method based on topic modeling. We ran 500 iterations of the model with the default parameters. 
\textit{Higher-Order Skip-Gram (HOSG)} by~\citet{Cotterell:17} generalizes the Skip-Gram model \cite{Mikolov:13} by extending it from word-context co-occurrence matrices to tensors factorized with a polyadic decomposition. In our case, this tensor consisted of SVO triple counts. We trained three vector arrays (for subjects, verbs and objects) on the 108,073 SVO triples from the \textit{FrameNet} corpus, using the  implementation by the authors. Training was performed with 5 negative samples, 300-dimensional vectors, and 10 epochs. We constructed an embedding of a triple by concatenating embeddings for subjects, verbs, and objects, and clustered them using $k$-means with the number of clusters set to 10,000 (this value provided the best performance).
\textit{NOAC}~\cite{Egurnov:17} is an extension of the Object Attribute Condition (OAC) triclustering algorithm~\cite{Ignatov:15} to numerically weighted triples. This incremental algorithm searches for dense regions in triadic data. A minimum density of 0.25 led to the best results.
In the \textit{Triadic} baselines, independent word embeddings of subject, object, and verb are concatenated and then clustered using a \textit{hard clustering algorithm}: $k$-means, spectral clustering, or CW.

We tested various hyper-parameters of each of these algorithms and report the best results overall per clustering algorithm. Two trivial baselines are \textit{Singletons} that creates a single cluster per instance and \textit{Whole} that creates one cluster for all elements.

\begin{table*}[t]
\resizebox{1.0\linewidth}{!}{
\centering
\begin{tabular}{l*{4}{|*{3}{r}}}
& \multicolumn{3}{c|}{\textbf{Verb}}
& \multicolumn{3}{c|}{\textbf{Subject}}
& \multicolumn{3}{c|}{\textbf{Object}}
& \multicolumn{3}{c}{\textbf{Frame}} \\\cmidrule{2-13}
\textbf{Method}  & \textbf{nmPU} & \textbf{niPU} & \textbf{F\textsubscript{1}}
                 & \textbf{nmPU} & \textbf{niPU} & \textbf{F\textsubscript{1}}
                 & \textbf{nmPU} & \textbf{niPU} & \textbf{F\textsubscript{1}}
                 & \textbf{nmPU} & \textbf{niPU} & \textbf{F\textsubscript{1}} \\\toprule
Triframes {\watset} & 42.84 & 88.35 & \textbf{57.70} & 54.22 & 81.40 & 65.09 & 53.04 & 83.25 & 64.80 & 55.19 & 60.81 & \textbf{57.87} \\
HOSG \cite{Cotterell:17} & 44.41 & 68.43 & 53.86 & 52.84 & 74.53 & 61.83 & 54.73 & 74.05 &    62.94 & 55.74 & 50.45 & 52.96 \\
NOAC \cite{Egurnov:17} & 20.73 & 88.38 & 33.58 & 57.00 & 80.11 & \textbf{66.61} & 57.32 & 81.13 & \textbf{67.18} & 44.01 & 63.21 & 51.89 \\
Triadic Spectral    & 49.62 & 24.90 & 33.15 & 50.07 & 41.07 & 45.13 & 50.50 & 41.82 & 45.75 & 52.05 & 28.60 & 36.91 \\
Triadic $k$-Means   & \textbf{63.87} & 23.16 & 33.99 & \textbf{63.15} & 38.20 & 47.60 & \textbf{63.98} & 37.43 & 47.23 & \textbf{63.64} & 24.11 & 34.97 \\
LDA-Frames \cite{Materna:13} & 26.11 & 66.92 & 37.56 & 17.28 & 83.26 & 28.62 & 20.80 & 90.33 & 33.81 & 18.80 & 71.17 & 29.75 \\
Triframes CW        &  7.75 &  6.48 &  7.06 &  3.70 & 14.07 &  5.86 & 51.91 & 76.92 & 61.99 & 21.67 & 26.50 & 23.84 \\\midrule
Singletons          &  0.00 & 25.23 &  0.00 &  0.00 & 25.68 &  0.00 &  0.00 & 20.80 &  0.00 & 32.34 & 22.15 & 26.29 \\
Whole               &  3.62 & \textbf{100.0} &  6.98 &  2.41 & \textbf{98.41} &  4.70 &  2.38 & \textbf{100.0} &  4.64 &  2.63 & \textbf{99.55} &  5.12 \\
\end{tabular}
}
\caption{\label{tab:frames}Frame evaluation results on the triples from the FrameNet 1.7 corpus~\cite{Baker:98}. The results are sorted by the descending order of the Frame F\textsubscript{1}-score. Best results are boldfaced.}
\vspace*{-.7em}
\end{table*}

\section{Results}\label{sec:results}

We perform two experiments to evaluate our approach: (1) a frame induction experiment on the FrameNet annotated corpus by~\citet{Bauer:12}; (2) the polysemous verb clustering experiment on the dataset by~\citet{Korhonen:03}. The first is based on the newly introduced frame induction evaluation schema (cf. Section~\ref{sec:evaluation}). The second one evaluates the quality of verb clusters only on a standard dataset from prior work.

\begin{figure}[t]
  \centering
  \includegraphics[width=\columnwidth]{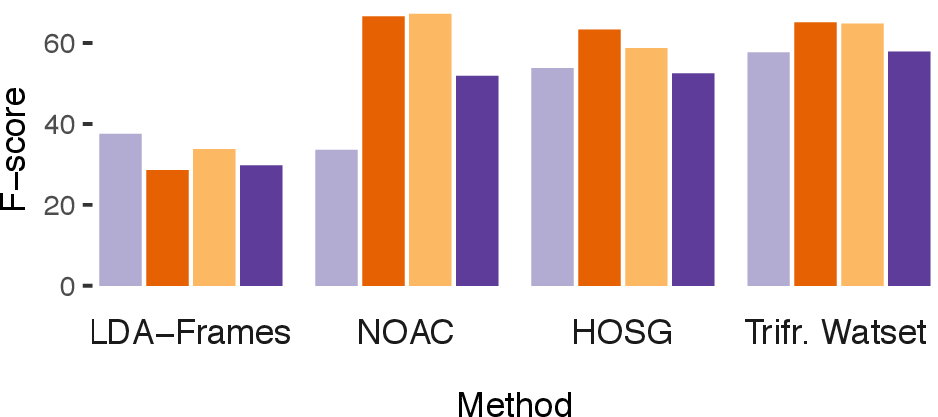} 
  \vspace{-1.5em}
  \caption{\label{fig:frames}F\textsubscript{1}-scores for \legend{ggplotverb}\,verbs, \legend{ggplotsubject}\,subjects, \legend{ggplotobject}\,objects, \legend{ggplotframe}\,frames corresponding to \tablename~\ref{tab:frames}.}
\end{figure}

\paragraph{Frame Induction Experiment.} In \tablename~\ref{tab:frames} and \figurename~\ref{fig:frames}, the results of the experiment are presented. Triframes based on {\watset} clustering outperformed the other methods on both Verb F\textsubscript{1} and overall Frame F\textsubscript{1}. The \textit{HOSG}-based clustering proved to be the most competitive baseline, yielding decent scores according to all four measures. The \textit{NOAC} approach captured the frame grouping of slot fillers well but failed to establish good verb clusters. Note that \textit{NOAC} and \textit{HOSG} use only the graph of syntactic triples and do not rely on pre-trained word embeddings. This suggests a high complementarity of signals based on distributional similarity and global structure of the triple graph. Finally, the simpler \textit{Triadic} baselines relying on hard clustering algorithms showed low performance, similar to that of \textit{LDA-Frames}, justifying the more elaborate {\watset} method.

While triples are intuitively less ambiguous than words, still some frequent and generic triples like $(\text{she}, \text{make}, \text{it})$ can act as hubs in the graph, making it difficult to split it into semantically plausible clusters. The poor results of the Chinese Whispers hard clustering algorithm illustrate this. Since the hubs are ambiguous, i.e., can belong to multiple clusters, the use of the {\watset} fuzzy clustering algorithm that splits the hubs by disambiguating them leads to the best results (see Table~\ref{tab:frames}). 

\paragraph{Verb Clustering Experiment.} \tablename~\ref{tab:verbs} presents results on the second dataset for the best models identified on the first dataset. The \textit{LDA-Frames} yielded the best results with our approach performing comparably in terms of the F\textsubscript{1}-score. We attribute the low performance of the Triframes method based on CW clustering to its hard partitioning output, whereas the evaluation dataset contains fuzzy clusters. Different rankings also suggest that frame induction cannot simply be treated as a verb clustering and requires a separate task.

\begin{table}[t]
\footnotesize
\centering
\begin{tabular}{l*{3}{r}}
\textbf{Method}     & \textbf{nmPU} &  \textbf{niPU} & \textbf{F\textsubscript{1}} \\\toprule
LDA-Frames          & \textbf{52.60} & 45.84 & \textbf{48.98} \\
Triframes {\watset} & 40.05 & 62.09 & 48.69 \\
NOAC                & 37.19 & 64.09 & 47.07 \\
HOSG                & 38.22 & 43.76 & 40.80 \\ 
Triadic Spectral    & 35.76 & 38.96 & 36.86 \\
Triadic $k$-Means   & 52.22 & 27.43 & 35.96 \\
Triframes CW        & 18.05 & 12.72 & 14.92 \\\midrule
Whole               & 24.14 & \textbf{79.09} & 36.99 \\
Singletons          &  0.00 & 27.21 &  0.00 \\
\end{tabular}
\caption{\label{tab:verbs}Evaluation results  on the dataset of polysemous verb classes by \citet{Korhonen:03}.}
\vspace{-1em}
\end{table}

\section{Conclusion}

In this paper, we presented the first application of \textit{triclustering} for unsupervised \textit{frame induction}. We designed a dataset based on the FrameNet and SVO triples to enable fair corpus-independent evaluations of frame induction algorithms. We tested several triclustering methods as the baselines and proposed a new graph-based triclustering algorithm that yields state-of-the-art results. 
A promising direction for future work is using the induced frames in applications, such as Information Extraction and Question Answering. 

Additional illustrations and examples of extracted frames are available in the supplementary materials. The source code and the data are available online under a permissive license.\footnote{\url{https://github.com/uhh-lt/triframes}}


\section*{Acknowledgments}

We acknowledge the support of  DFG  under the ``JOIN-T'' and ``ACQuA'' projects and thank three anonymous reviewers for their helpful comments. Furthermore, we thank Dmitry Egurnov, Dmitry Ignatov, and Dmitry Gnatyshak for help in operating the NOAC method using the multimodal clustering toolbox. Besides, we are grateful to Ryan Cotterell and Adam Poliak for a discussion and an implementation of the HOSG method. Finally, we thank Bonaventura Coppolla for discussions and preliminary work on graph-based frame induction.


\bibliography{triframes.acl2018}
\bibliographystyle{acl_natbib}

\end{document}